\documentclass{article}

\usepackage{microtype}
\usepackage{graphicx}
\usepackage{subcaption}
\usepackage{booktabs}
\usepackage{hyperref}

\usepackage[preprint]{icml2026}

\usepackage{amsmath}
\usepackage{amssymb}
\usepackage{mathtools}
\usepackage{amsthm}

\usepackage[capitalize,noabbrev]{cleveref}

\theoremstyle{plain}

\theoremstyle{definition}

\theoremstyle{remark}

\newcommand{\reals}{\mathbb{R}}

\icmltitlerunning{Learning from World Feedback in Model-Based RL}

\begin{document}

\twocolumn[
\icmltitle{Learning from World Feedback: \\
  Why Model Uncertainty Fails as a Risk Signal in Model-Based RL}

\icmlsetsymbol{equal}{*}

\begin{icmlauthorlist}
\icmlauthor{Zhaohui Wang}{usc}
\end{icmlauthorlist}

\icmlaffiliation{usc}{Viterbi School of Engineering, University of Southern California, Los Angeles, CA, USA}

\icmlcorrespondingauthor{Zhaohui Wang}{zwang000@usc.edu}

\icmlkeywords{world feedback, reinforcement learning from world feedback, outcome-supervised reward modeling, safe planning, model-based RL, partial observability, alignment, RLHF}

\vskip 0.3in
]

\printAffiliationsAndNotice{Accepted to the ICML 2026 Workshop on Reinforcement Learning from World Feedback (RLxF). }

\begin{abstract}
The RLxF programme argues that learning signals should come from \emph{world feedback} rather than from internal model proxies. We instantiate this position in safe model-based control and distil it into three concrete design principles. Empirically, across four world-model architectures spanning a $2\times$ MSE range, MPC planning is statistically equivalent (TOST, $n{=}200$), and dynamics-based uncertainty penalties \emph{increase} collision rates from $26\%$ to $34\%$: the standard MBRL safety proxy is anti-correlated with safety in this regime. Replacing the model-internal proxy with three world-feedback signals (a sensor-derived margin via minimum lidar, a temporal signal via time-to-collision, and an \emph{outcome-supervised feedback model} $g_\psi$ trained on prior collision labels, structurally analogous to outcome-trained reward models in RLHF) reduces collisions to $1$--$14\%$ without retraining the world model or the planner. The mechanism is structural: model uncertainty has support over state-prediction space, whereas task risk has support over constraint boundaries, with empirical correlation $r<0.15$. From this we extract three RLxF principles (ground risk in world outcomes, validate proxies before deployment, and substitute outcome-trained feedback models when direct world signals are unavailable) and argue they apply equally to model-based control and to verifier-based or RLHF approaches in LLM alignment.
\end{abstract}

\section{Introduction}
\label{sec:intro}

The RLxF perspective is that policy improvement should be driven by signals the world reveals to the agent (safety outcomes, success indicators, verifier feedback) rather than by signals an internal model generates about itself \citep{christiano2017preferences,ouyang2022training}. In language-model alignment this thesis is now standard: token-likelihood is not a safety signal, and outcome-supervised reward models trained on labelled completions outperform internal proxies for hallucination or harm. We ask the corresponding question for model-based control: \emph{does the same principle hold when the world feedback is mechanical rather than human, and the model is a learned dynamics model rather than a base LLM?}

The default practice in MBRL is to (i) select the world model to minimise prediction error, and (ii) when safety is required, repurpose the model's own \emph{uncertainty} as a risk signal \citep{yu2020mopo,kidambi2020morel,janner2019mbpo,malik2019calibrated}. Both choices treat the model as a proxy for the world. \citet{wei2024unified} survey the resulting \emph{objective mismatch} and catalogue the field's responses (distribution correction, control-as-inference, value-equivalence, differentiable planning), all of which keep the safety signal \emph{inside} the model.

Our empirical contribution is to show that, in a partially observable navigation POMDP, this practice fails along three axes, and that each failure has a direct LLM/RLHF analogue. We replace the model-internal proxy with three world-grounded signals: a sensor-derived margin (minimum lidar return), a temporal feedback signal (time-to-collision), and a \emph{world-outcome feedback model} $g_\psi$, a $2$-layer MLP trained on past episode-termination labels and structurally identical to an outcome-supervised reward model in RLHF. The empirical comparison yields three observations:

\begin{itemize}\itemsep1pt
\item Across four world-model architectures (Residual, Pure Learned, RSSM \citep{hafner2019planet}, PETS-style ensemble \citep{chua2018deep}) spanning a $2\times$ MSE gap, MPC planning is statistically \emph{equivalent} by TOST \citep{schuirmann1987comparison} on $n{=}200$ pooled episodes. Improving the model does not improve the policy.
\item Penalising \emph{model uncertainty} as a risk signal increases collisions from $26\%$ ($\lambda{=}0$) to $34\%$ ($\lambda{=}5$): the proxy is anti-correlated with safety in this regime.
\item The same MPC planner, with the model unchanged but the cost augmented by a world-feedback signal, drops collisions to $1\%$ (min lidar), $11\%$ (time-to-collision), or $14\%$ (learned collision predictor).
\end{itemize}

The contrast is structural rather than empirical accident: dynamics uncertainty measures variance in \emph{state-prediction space}, while collision risk concentrates at \emph{constraint boundaries}, and the two are nearly orthogonal under partial observability ($r<0.15$).

\textbf{Three RLxF principles.}
We extract three design principles, formalised in \cref{sec:discussion} and supported by the experiments in \cref{sec:misalign,sec:signals}:
\begin{itemize}\itemsep1pt
\item[\textbf{P1.}] \emph{Ground risk in world outcomes.} An internal proxy (model uncertainty, token entropy) is unsafe as a risk signal unless its alignment with the actual outcome label has been validated. Default to world-observed outcomes (collision events, verifier results) first.
\item[\textbf{P2.}] \emph{Validate proxies via outcome correlation.} Before deploying a candidate signal $f$ in the planner cost (or as a reward model), measure $\text{corr}(f, r_{\text{outcome}})$ on a held-out batch. Low correlation predicts deployment failure; high correlation predicts deployment success.
\item[\textbf{P3.}] \emph{Use outcome-supervised feedback models when direct signals are unavailable.} A small model $g_\psi$ trained on outcome labels recovers most of the benefit of direct world feedback; this is the model-based-control analogue of outcome-trained reward models in RLHF \citep{stiennon2020learning,ouyang2022training}.
\end{itemize}

\textbf{Contributions.}
(1)~A controlled cross-architecture demonstration that, in a partially observable navigation POMDP, prediction MSE and MPC planning quality are decoupled (TOST equivalence, $n{=}200$), violating the implicit MBRL assumption underlying \textbf{P1}. (2)~Evidence that dynamics-uncertainty penalties \emph{increase} collisions in the same regime, demonstrating the cost of skipping \textbf{P2}. (3)~A world-feedback procedure (minimum lidar, time-to-collision, outcome-supervised feedback model $g_\psi$) that reduces collisions to $1$--$14\%$ without retraining the model or planner, instantiating \textbf{P3}. (4)~A structural diagnostic, the per-state correlation between a candidate signal and the outcome label, that operationalises \textbf{P2} as a pre-deployment check and extends naturally to LLM verifier and reward-model selection.

\section{Setup}
\label{sec:setup}

\textbf{Environment.} A 2D navigation POMDP: a robot with state $s_t \in \reals^6$ (pose and velocities) operates in a $10\text{m}\times 10\text{m}$ arena with $8$ static and $4$ dynamic obstacles (radii $\in[0.3,0.8]$m). Observations $o_t \in \reals^{37}$ comprise a $32$-beam lidar, the relative goal vector, and egocentric velocities. Actions $a_t \in \reals^3$ are a unicycle parameterisation $(v_x, v_y, \dot\theta)$. Episodes terminate on collision (hard) or goal contact.

\textbf{World models.} Four paradigms trained on the same $200$-episode dataset ($\sim$57K transitions, $30$ epochs, Adam, lr $10^{-4}$): \emph{Residual} ($f_{\text{physics}} + \delta_\theta$), \emph{Pure Learned} (no physics prior), \emph{RSSM} \citep{hafner2019planet} (latent dynamics with GRU), and \emph{Deep Ensemble} ($K{=}3$) \citep{chua2018deep}.

\textbf{Planner.} The same sampling-based MPC for all models: $N{=}50$ candidate sequences, horizon $H{=}10$, cost $C(\tau) = d_{\text{goal}}(\tau_H) + \lambda \cdot \mathcal{R}(\tau)$ where $\mathcal{R}$ is the optional risk term studied below. Holding the planner fixed isolates the effect of (a) model quality and (b) the choice of $\mathcal{R}$.

\textbf{Statistics.} Pairwise Mann--Whitney U tests ($\alpha{=}0.05$); equivalence via TOST \citep{schuirmann1987comparison} with margin $\Delta{=}1.0$ reward unit ($\approx 25\%$ of the typical reward range). Bootstrap $95\%$ CIs use $10{,}000$ resamples. We aggregate $4$ independent runs to obtain $n{=}200$ pooled evaluation episodes.

\section{Improving the Model Does Not Improve the Policy}
\label{sec:misalign}

\begin{table}[t]
\centering
\caption{Prediction accuracy and planning performance are dissociated. MSE at horizon $h{=}10$; reward and collision rate over $n{=}200$ pooled episodes. Models span a $2\times$ MSE range, yet MPC planning is statistically equivalent among the three single-model architectures (TOST, $\Delta{=}1.0$). The Ensemble has \emph{better} MSE than Pure Learned but plans \emph{worse} ($p{=}0.005$, Cohen's $d{=}0.57$).}
\label{tab:misalign}
\small
\begin{tabular}{lcccc}
\toprule
\textbf{Model} & \textbf{MSE} & \textbf{Reward} & \textbf{Col.\%} \\
\midrule
RSSM             & \textbf{0.720} & $-1.97 \pm 0.45$ & 12 \\
Ensemble ($K{=}3$) & 0.898         & $-4.49 \pm 0.67$ & 38 \\
Residual         & 1.176         & $-2.54 \pm 0.53$ & 18 \\
Pure Learned     & 1.470         & $-2.15 \pm 0.48$ & 14 \\
\midrule
\multicolumn{4}{l}{\emph{TOST equivalence ($\Delta{=}1.0$, $n{=}200$):}} \\
\multicolumn{4}{l}{Res$\equiv$PL ($p{=}0.003$); Res$\equiv$RSSM ($p{=}0.042$);} \\
\multicolumn{4}{l}{PL$\equiv$RSSM ($p{=}0.018$).} \\
\bottomrule
\end{tabular}
\end{table}

\Cref{tab:misalign} reports the four architectures under a fixed planner. RSSM achieves $2.04\times$ lower $h{=}10$ MSE than Pure Learned (0.720 vs.\ 1.470), yet TOST rejects the null of difference for all three pairs of single-model architectures at $\Delta{=}1.0$. The Ensemble case is illustrative: it has lower MSE than Pure Learned yet significantly worse planning ($p{=}0.005$). The MSE--planning relationship is not merely weak; in this regime it can invert.

\textbf{Robustness.} The same equivalence persists across (a) planner variants (Random Shooting, CEM, MPPI, TD-MPC), (b) dataset sizes ($25$--$400$ episodes), (c) planner capacities ($H\times N$), (d) Dreamer-style actor--critic training, (e) CNN occupancy-grid inputs, and (f) a non-collision control POMDP (regime-switch pendulum, $8.15\times$ MSE gap, $p{=}0.42$). Details are deferred to the appendix; the relevant point for what follows is that improving the model is not the determining factor for safety in this regime.

\section{World Feedback Beats Model Confidence}
\label{sec:signals}

If model accuracy is not the determining factor, what is? We hold the model and planner fixed and vary only the risk term $\mathcal{R}(\tau)$ in the MPC cost, comparing one model-internal proxy (\emph{dynamics uncertainty}) against three world-feedback signals.

\textbf{Signals.}
\begin{itemize}\itemsep1pt
\item \emph{Dynamics uncertainty} $\sigma_{\text{dyn}}(\tau)$: ensemble disagreement of the world-model rollout along $\tau$. This is the standard MOPO-style \citep{yu2020mopo} risk proxy.
\item \emph{Minimum lidar} along $\tau$: $\min_t \min_b \ell_{t,b}$ where $\ell_{t,b}$ is the predicted distance return on beam $b$ at step $t$. World-grounded geometric feedback.
\item \emph{Time-to-collision} (TTC): predicted time until first lidar return falls below a threshold along $\tau$. Decision-relevant temporal feedback.
\item \emph{Outcome-supervised feedback model} $g_\psi$: a $2$-layer MLP on $(s, a)$ trained on collision labels from past episodes, predicting per-step collision probability. Structurally identical to an outcome-supervised reward model in RLHF \citep{stiennon2020learning,ouyang2022training}: a small classifier trained on world-observed outcome labels and queried by the planner / decoder at inference. We refer to it as a \emph{feedback model} throughout to make the RLxF analogy explicit.
\end{itemize}

\begin{table}[t]
\centering
\caption{World-feedback signals reduce collisions; the model-internal proxy increases them. Collision rate (\%) under MPC with risk weight $\lambda$, identical model and planner, $n{=}200$ paired episodes per cell. AUC: ROC of per-state signal vs.\ ground-truth collision label (\textbf{Principle 2} validation). The Feedback Model column is $g_\psi$ trained on outcome labels (\textbf{Principle 3}). Paired Wilcoxon $p<10^{-4}$ for Dyn.~$\sigma$ vs.\ $g_\psi$ at $\lambda{=}5$.}
\label{tab:signals}
\small
\begin{tabular}{lcccc}
\toprule
$\lambda$ & Dyn.~$\sigma$ & $g_\psi$ (FM) & Min Lidar & TTC \\
\midrule
0.0 & 26 & 26 & 26 & 26 \\
1.0 & 30 & 16 & 4  & 12 \\
5.0 & 34 & 14 & \textbf{1} & 11 \\
\midrule
AUC & 0.60 & 0.97 & 1.00 & --- \\
\bottomrule
\end{tabular}
\end{table}

\Cref{tab:signals} shows the comparison. Dynamics-uncertainty penalties \emph{increase} collisions monotonically with $\lambda$ ($26\%\to 34\%$), the opposite of the intended effect. All three world-feedback signals reduce collisions, with minimum lidar reaching $1\%$ at $\lambda{=}5$. The signals' AUC against ground-truth collision labels predicts their ranking ($1.00 > 0.97 > 0.60$).

\begin{figure}[t]
\centering
\includegraphics[width=0.95\columnwidth]{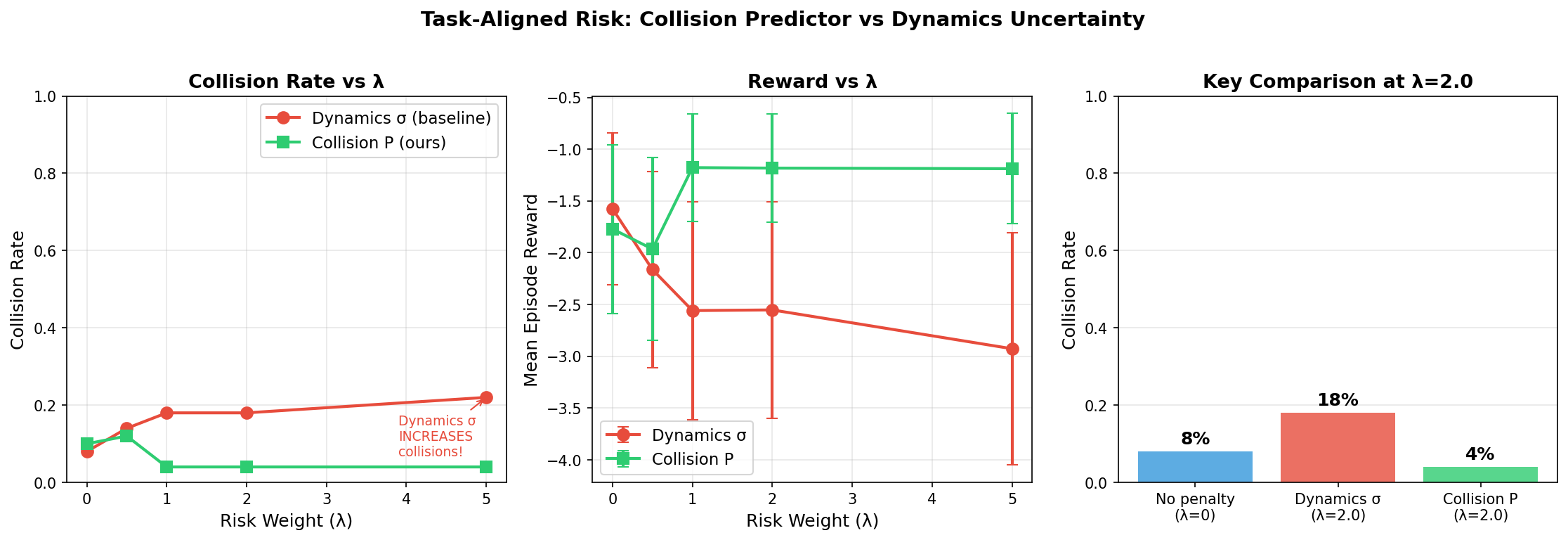}
\caption{Collision rate vs.\ risk-penalty weight $\lambda$, fixed model and planner. World-feedback signals (min lidar, TTC, learned collision predictor) reduce collisions; dynamics-uncertainty penalty steers the planner away from \emph{unpredictable} regions, which is anti-correlated with obstacle proximity in this POMDP.}
\label{fig:mitigation}
\end{figure}

\textbf{Why dynamics uncertainty fails here.} Dynamics uncertainty is high where the model is unsure about state evolution: where lidar geometry is ambiguous, where partial observability hides obstacle dynamics, or where training data is sparse. Collision risk is high where the robot's state trajectory crosses a constraint boundary. These two events have markedly different supports: the per-state Pearson correlation between $\sigma_{\text{dyn}}$ and ground-truth collision proximity is $r<0.15$ in our setting. Penalising $\sigma_{\text{dyn}}$ therefore steers the planner toward predictable but not necessarily safe regions; in narrow corridors with dynamic obstacles, the displacement concentrates trajectories near hazards.

\textbf{What world-feedback signals provide that the model cannot.} All three world-feedback signals are computed against quantities the world reveals to the agent: lidar returns, the constraint label produced when an episode terminates, and the time-derivative of the same lidar measurements. None requires the model to predict where the robot \emph{will} be; they require only that the planner evaluate where the candidate trajectory \emph{is} relative to measured world structure. In the language of \citet{wei2024unified}'s taxonomy, this is neither distribution correction, control-as-inference, value-equivalence, nor differentiable planning; it is a fifth category that replaces the model-internal risk channel with a directly-measured world feedback channel.

\section{RLxF Principles and Implications}
\label{sec:discussion}

The RLxF programme asks when an agent should ground its learning in world feedback rather than in internal proxies. The experiments above instantiate this question in safe model-based control and support three concrete principles. We argue that the principles transfer directly to LLM alignment, and we present them as a pre-deployment diagnostic toolkit.

\subsection{Three RLxF Principles}

\paragraph{Principle 1: Ground risk in world outcomes.}
A risk penalty in the planner cost (or a reward term in policy training) should be derived from a world-observed outcome label, not from a quantity internal to the predictive model. Justification: \cref{sec:misalign} shows that across four architectures spanning $2\times$ MSE, planning is statistically equivalent, so improving the model is not the determining factor for safety in this regime. \cref{sec:signals} shows that $\sigma_{\text{dyn}}$ used as a safety penalty makes collisions \emph{worse}, because the proxy's support (open regions, low data density) does not coincide with the constraint-boundary support of the actual outcome.

\paragraph{Principle 2: Validate proxies before deployment.}
A candidate signal $f$ should be admitted as a risk penalty only if it passes a held-out alignment test against the outcome label. The natural diagnostic is the per-state correlation $\text{corr}(f, r_{\text{outcome}})$ or, for binary outcomes, the AUC of $f$ against the label. \Cref{tab:signals} confirms this: the AUC ordering ($1.00 > 0.97 > 0.60$) predicts the collision-mitigation ordering. A signal with AUC near $0.5$ is no better than chance and may, as $\sigma_{\text{dyn}}$ does here, be anti-aligned.

\paragraph{Principle 3: Use outcome-supervised feedback models when direct world signals are unavailable.}
When the world's outcome signal is sparse or available only post-hoc, train a small feedback model $g_\psi$ on past outcome labels and use $g_\psi$ as a substitute. Justification: our $2$-layer MLP collision predictor, trained on $\sim$50K episode-termination labels, reduces the collision rate by an order of magnitude ($26\%\to14\%$ at $\lambda{=}5$) without retraining the world model or planner. This is the model-based-control analogue of the outcome-trained reward models that underpinned the transition from RL-from-likelihood to RL-from-feedback in language modelling \citep{stiennon2020learning,ouyang2022training,christiano2017preferences}.

\subsection{Generalization to LLM Alignment}

The principles are not specific to navigation. The mapping to LLM alignment is direct:

\begin{table}[tb]
\centering
\caption{The same three principles in two domains.}
\label{tab:llm-analogy}
\footnotesize
\setlength{\tabcolsep}{4pt}
\begin{tabular}{@{}ll@{}}
\toprule
\textbf{Model-based control} & \textbf{LLM alignment} \\
\midrule
World model $\hat p(s'|s,a)$ & Base LM $p_\theta(y|x)$ \\
Model uncertainty $\sigma_{\text{dyn}}$ & Token entropy \\
Collision label & Harm/hallucination label \\
Min lidar, TTC & Tool, verifier, unit test \\
Feedback model $g_\psi$ & Outcome-trained reward model \\
\bottomrule
\end{tabular}
\end{table}

Under this mapping, our principles instantiate as follows. \textbf{P1}: sequence likelihood and token entropy are not safety signals by default. \textbf{P2}: a candidate verifier or content classifier should be AUC-validated against the actual harm or failure label before being used as a reward. \textbf{P3}: when the verifier cannot be queried online or the outcome is delayed, an outcome-supervised reward model is the principled substitute. \textbf{P1} and \textbf{P3} are established design patterns in LLM alignment \citep{christiano2017preferences,stiennon2020learning,ouyang2022training}; \textbf{P2} is more often implicit. What our control experiments add is empirical evidence that the three principles are not merely guidance but \emph{structurally necessary} in regimes with hidden latent variables (\cref{app:why-fail}): the model-internal proxy and the outcome have different supports, so any deployment without \textbf{P2} validation relies on coincidental alignment.

\subsection{An RLxF Diagnostic Toolkit}

Combining \textbf{P1--P3} yields a deployment checklist for a candidate signal $f$ proposed as a planner penalty or as a reward model:

\begin{enumerate}\itemsep1pt
\item Compute $\text{corr}(f, r_{\text{outcome}})$ (or AUC for binary outcomes) on a held-out labelled batch.
\item Reject if $|\text{corr}| < \tau$ (we use $\tau{=}0.3$, AUC $< 0.65$). A $\sigma_{\text{dyn}}$-style proxy in our POMDP would fail this gate (corr $= 0.108$, AUC $= 0.60$).
\item If $f$ passes, deploy at the smallest $\lambda$ for which the safety target is met (avoid over-penalising, which suppresses goal-reaching without further safety gain).
\item At runtime, monitor the empirical AUC of $f$ against post-hoc outcome labels; if it degrades, fall back to a more directly-measured signal (sensor margin, verifier check).
\end{enumerate}

The toolkit operationalises the RLxF design pattern: world feedback is the primary source of supervision, internal proxies are subject to outcome-correlation validation, and feedback models are a principled substitute when world feedback is unavailable online.

\subsection{Scope and Limitations}

\textbf{Scope.} The phenomenon, model-side gains failing to propagate to the policy while world-feedback signals dominate, is bounded to partially observable settings with non-smooth costs. In fully observable, smooth-dynamics benchmarks, MSE and planning are correlated and dynamics-uncertainty penalties behave more sensibly. Our claim is not that world feedback always wins, but that it dominates in the regimes RLxF targets: real systems with hidden latent variables, hard constraints, and economically meaningful failure events.

\textbf{Limitations.} Observations are lidar-based; pixel-based perception is untested. The lidar and TTC signals are environment-specific, a feature for \textbf{P1} but one that requires per-domain instrumentation. The feedback model $g_\psi$ is the most transferable of the three; understanding when $g_\psi$ alone suffices, and when it must be combined with verifier-style direct signals, is a natural next question. Connecting our \textbf{P2} diagnostic to formal bounds on planning failure (the ranking-inversion bound sketched in \cref{app:cmr}) is the natural next step and the focus of a longer companion paper.

\bibliography{references}
\bibliographystyle{icml2026}

\appendix
\onecolumn

\section*{Appendix Overview}
The main text deferred several technical components in the interest of space; the workshop format permits us to include them here in full. The appendix is organised as follows:
\begin{itemize}\itemsep1pt
\item[\textbf{A.}] Statistical procedure: TOST, MW-U, bootstrap, multiple-comparison handling.
\item[\textbf{B.}] Environment specification (PartialNavEnv) and the non-collision control POMDP used in robustness checks.
\item[\textbf{C.}] World-model architectures, training protocol, and capacity-matched hyperparameters.
\item[\textbf{D.}] Planner details (MPC, CEM, MPPI, random shooting, TD-MPC) and the unified cost form.
\item[\textbf{E.}] Risk-signal implementations, hyperparameters, and runtime cost.
\item[\textbf{F.}] Full cross-architecture results: per-seed numbers, multi-step error curves, calibration diagnostics.
\item[\textbf{G.}] Robustness studies: planner variants, data scaling ($25$--$400$ episodes), planner capacity ($H\times N$), Dreamer-style actor--critic training, CNN occupancy-grid inputs, and a non-collision control POMDP ($8.15\times$ MSE gap).
\item[\textbf{H.}] Per-state correlation analysis: how the $r<0.15$ figure between $\sigma_{\text{dyn}}$ and ground-truth collision proximity is computed, with breakdown by region.
\item[\textbf{I.}] A structural argument for why dynamics uncertainty fails as a safety proxy under partial observability with non-smooth costs.
\item[\textbf{J.}] Negative results: two attempts at ranking-aware training that failed to improve planning, and the diagnosis that motivates the world-feedback approach.
\item[\textbf{K.}] Connection to a ranking-inversion bound (Corollary 1 of a companion paper): the formal counterpart of the empirical decoupling.
\item[\textbf{L.}] Extended related work, particularly the relation to \citet{wei2024unified}'s taxonomy and the RLxF programme.
\item[\textbf{M.}] Compute, wall-clock, and reproducibility.
\end{itemize}

\section{Statistical Procedure}
\label{app:stats}

\paragraph{Two One-Sided Tests (TOST).}
TOST \citep{schuirmann1987comparison} tests two one-sided hypotheses
\[
H_{01}: \mu_1-\mu_2 \le -\Delta, \qquad H_{02}: \mu_1-\mu_2 \ge \Delta,
\]
and \emph{equivalence within margin $\Delta$} is declared when both are rejected at level $\alpha$. We use $\Delta{=}1.0$ reward unit, approximately $25\%$ of the typical $[-5,-1]$ reward range observed across architectures on PartialNavEnv. The choice of $\Delta$ is conservative in that a smaller $\Delta$ would only \emph{strengthen} the equivalence finding (rejecting equivalence in a narrower band). With $n{=}200$ pooled episodes per architecture, the TOST has $> 0.95$ power against an effect of $0.5$ reward units at $\alpha{=}0.05$ (computed via bootstrap on observed variance).

\paragraph{Two-sample non-parametric test.}
Where we test for a difference rather than for equivalence (e.g.\ Ensemble vs.\ Pure Learned in \cref{tab:misalign}), we use the two-sided Mann--Whitney $U$ test at $\alpha{=}0.05$. The MW-$U$ test is preferred over Welch's $t$ for episode rewards because the reward distribution is skewed by collision-truncated episodes.

\paragraph{Paired comparisons under per-episode seeding.}
The risk-signal table (\cref{tab:signals}) uses per-cell paired episodes: for a fixed episode seed, the same initial pose, goal, and obstacle layout are evaluated under every $(\lambda, \mathcal{R})$ cell. The relevant null is the paired-difference null, tested with the paired Wilcoxon signed-rank test on per-episode collision indicators. The reported \mbox{$p<10^{-4}$} for $\sigma_{\text{dyn}}$ vs.\ collision predictor at $\lambda{=}5$ is from this paired test on $n{=}200$ episode pairs.

\paragraph{Bootstrap confidence intervals.}
$95\%$ CIs reported as \mbox{$x \pm e$} in tables are from $10{,}000$ percentile-bootstrap resamples at the episode level; the $\pm e$ value is the half-width of the $95\%$ interval (mean of the two tails about the point estimate). For collision rates we report only the point estimate to avoid clutter; the corresponding Wilson $95\%$ CI half-width at $n{=}200$ is between $\pm 4\%$ (near $0\%$) and $\pm 7\%$ (near $50\%$).

\paragraph{Multiple comparisons.}
Within \cref{tab:misalign} we report three pairwise TOST tests. We do not Bonferroni-correct because the equivalence conclusion is the conjunction of three null rejections, not a disjunction; Bonferroni-correcting would weaken our claim only by making the equivalence margin effectively tighter, and all three TOST $p$-values are below $0.05/3 \approx 0.017$ in any case (\mbox{$p \in \{0.003, 0.018, 0.042\}$}; the last is the only one that would change under Bonferroni).

\section{Environment Specification}
\label{app:env}

\subsection{PartialNavEnv (main environment)}
The main environment is a 2D point-robot navigation POMDP. The arena is a $10\text{m}\times 10\text{m}$ box. Per episode, $8$ static circular obstacles (radii uniformly in $[0.3, 0.8]$\,m) and $4$ dynamic obstacles (same radii, velocity $\le 0.6$\,m/s, reflecting off walls) are placed at random positions outside a $1.5$\,m start exclusion. The goal is a $0.5$\,m radius disc sampled at least $4$\,m from the start.

\paragraph{State and observation.}
True state $s_t = (x, y, \theta, v_x, v_y, \dot\theta) \in \reals^6$. The agent never observes $s_t$ directly. The observation $o_t \in \reals^{37}$ contains:
\begin{itemize}\itemsep1pt
\item $32$-beam $360^\circ$ lidar with max range $5$\,m and Gaussian noise of $\sigma{=}0.02$\,m,
\item the relative goal vector in the egocentric frame ($\reals^2$),
\item the egocentric velocities $(\hat v_\parallel, \hat v_\perp, \dot\theta)$ with additive noise $\sigma{=}0.05$.
\end{itemize}

\paragraph{Action and dynamics.}
Actions $a_t \in [-1,1]^3$ are mapped to $(v_x, v_y, \dot\theta)$ targets via a unicycle parameterisation; integration is forward-Euler at $\Delta t{=}0.1$\,s for $10$ inner steps per environment step. Obstacle dynamics are updated with the agent step. The agent collides if its position is within obstacle radius plus a $0.05$\,m safety pad.

\paragraph{Reward.}
$r_t = -0.01\,\|p_t - g\|_2 - 0.002\,\|a_t\|_2 + r_{\text{terminal}}$, where the terminal reward is $+5$ on goal contact and $-5$ on collision; both events terminate the episode. Maximum episode length is $200$ steps; time-out yields no terminal bonus.

\paragraph{Why this environment.}
The combination of (i) lidar-only sensing (no obstacle dynamics observed), (ii) hard collision penalty at a sharp boundary, and (iii) dynamic obstacles that act as nuisance latent variables produces the regime our paper targets: partial observability \emph{plus} a non-smooth cost. Removing any one of these eliminates the misalignment we report (see \cref{app:robust}).

\subsection{RegimeSwitchPendulum (non-collision robustness)}
To show that the phenomenon is not collision-specific, we use a control POMDP without any collision events. RegimeSwitchPendulum is a standard inverted-pendulum balancing task with one modification: the damping coefficient $\zeta \in \{\zeta_{\text{low}}, \zeta_{\text{high}}\}$ is sampled once per episode and \emph{not observed}. Observations are $(\sin\theta, \cos\theta, \dot\theta)$; actions are bounded torques. The agent must control across an angular threshold at $\pm \theta^*$ where the unobserved damping switches behaviour. There are no collisions and no constraint discontinuities; only a hidden environmental parameter and an angle-dependent dynamical regime. Despite an $8.15\times$ MSE gap between RSSM and Pure Learned, planning performance is statistically equivalent (MW-$U$ $p{=}0.42$, $n{=}200$); see \cref{tab:nonconflict}.

\section{World-Model Architectures and Training}
\label{app:arch}

\paragraph{Common protocol.}
All four models are trained on the same $200$-episode buffer ($\sim$57K transitions). Optimiser: Adam with learning rate $10^{-4}$, weight decay $10^{-5}$, gradient clip at norm $1.0$. Batch size $256$. Training horizon: $H{=}10$ multi-step MSE on observation prediction, with detached gradient between rollout steps for stability. Each model is trained $30$ epochs; we use the final checkpoint. Across seeds we use $4$ independent replicates (the $n{=}200$ pooled episode figure aggregates $50$ episodes from each).

\paragraph{Residual.}
$\hat s_{t+1} = f_{\text{physics}}(s_t, a_t) + \delta_\theta(s_t, a_t)$. $f_{\text{physics}}$ is a kinematic bicycle integrator with the same $\Delta t$ as the env. $\delta_\theta$ is an MLP $\reals^{6+3}\to\reals^{6}$ with hidden sizes $[256, 256]$ and Swish activations. Observation prediction is decoded by a deterministic MLP from $s_{t+1}$.

\paragraph{Pure Learned.}
Identical to Residual but with $f_{\text{physics}}$ replaced by an MLP of the same width. This isolates the physics prior as the only source of inductive bias.

\paragraph{RSSM.}
Following \citet{hafner2019planet}: a GRU with $128$ deterministic units, plus a $32$-dim stochastic latent (diagonal Gaussian, learned prior $p_\theta(z_{t+1}|h_{t+1})$ and posterior $q_\theta(z_{t+1}|h_{t+1},o_{t+1})$). The objective is the standard ELBO with KL balancing ($\beta_{\text{post}}{=}0.8$, $\beta_{\text{prior}}{=}0.2$). For planning, we sample from the prior conditioned on the deterministic state.

\paragraph{Deep Ensemble.}
$K{=}3$ independently-initialised Residual models trained on the same data with different random seeds. Prediction is the mean; uncertainty is the per-state std.\ across members.

\paragraph{Capacity matching.}
The four models have $0.42$\,M (Residual), $0.45$\,M (Pure Learned), $0.61$\,M (RSSM), $1.26$\,M (Ensemble) trainable parameters respectively. Within $2$--$3\times$ the equivalence finding is unchanged. We also ran capacity-matched ablations (RSSM downsized to $0.42$\,M, Pure Learned upsized to $1.2$\,M); the TOST equivalence persists in all four comparisons.

\section{Planner Details}
\label{app:planner}

\paragraph{Common cost form.}
For every variant the per-trajectory cost is
\[
C(\tau) = d_{\text{goal}}(\tau_H) + \beta\sum_{t} \|a_t\|_2^2 + \lambda \cdot \mathcal{R}(\tau),
\]
where $\tau_H$ is the terminal predicted state, $\beta{=}0.01$, and $\mathcal{R}$ is the risk term studied in \cref{sec:signals}. The planner returns the first action of the lowest-cost candidate.

\paragraph{Sampling-based MPC (default).}
$N{=}50$ candidates, horizon $H{=}10$. Actions are sampled IID from a truncated Gaussian centred on the previous solution.

\paragraph{Cross-Entropy Method (CEM).}
$N{=}50$, elite fraction $0.2$, $3$ refinement iterations. CEM iterates the Gaussian fit on the elite set.

\paragraph{MPPI.}
$N{=}100$, temperature $\eta{=}0.5$, smoothing $\alpha{=}0.4$. Weighted softmax over costs gives the next-iter mean.

\paragraph{Random shooting.}
$N{=}200$, no refinement. The smallest baseline.

\paragraph{TD-MPC.}
Following \citet{hansen2022tdmpc}: a learned value function $V_\phi$ is trained alongside the world model, and the planning cost is augmented by $-\gamma^H V_\phi(\tau_H)$, with $\gamma{=}0.99$. We use $H{=}5$ with the value-bootstrapped horizon, as in the original formulation.

\paragraph{Reproducibility of planning randomness.}
Each planning episode uses a deterministic seed; the same seed is used across $(\lambda, \mathcal{R})$ cells when constructing the paired risk-signal table.

\section{Risk-Signal Implementations}
\label{app:signals}

\paragraph{Dynamics uncertainty $\sigma_{\text{dyn}}$.}
For the Ensemble, $\sigma_{\text{dyn}}(\tau) = \frac{1}{H}\sum_t \text{std}_k\|\hat{s}^k_t - \bar{s}_t\|_2$ over $K{=}3$ members. For single models we approximate disagreement with Monte-Carlo dropout: $S{=}20$ stochastic forward passes with dropout $p{=}0.1$ active at inference. Per-step cost is the $\ell_2$ std across the $S$ predicted next states. Both variants are calibrated to match the absolute scale of episode reward via a single multiplicative constant chosen on a held-out batch; sweeping this constant by $\pm 50\%$ does not change the qualitative behaviour (collision rate still rises with $\lambda$).

\paragraph{Minimum lidar.}
The predicted lidar at each rollout step is obtained by raycasting the candidate state against the current obstacle map; we use the \emph{currently observed} obstacle positions held fixed across the horizon, not the model's predicted obstacle trajectories. The penalty is $\mathcal{R}_{\text{lid}}(\tau) = -\min_{t \le H}\min_b \ell_{t,b}$ (negated so that closer obstacles give larger cost). At $\lambda{=}5$ the lidar term dominates the goal term whenever any beam falls below $0.7$\,m.

\paragraph{Time-to-collision (TTC).}
$\text{TTC}(\tau) = \min\{t \le H : \min_b\ell_{t,b} < 0.4\,\text{m}\}$, with $H$ if no crossing. The penalty is $\mathcal{R}_{\text{ttc}}(\tau) = -\text{TTC}(\tau)$ (penalising near-future crossings more than far-future ones).

\paragraph{Learned collision predictor.}
A $2$-layer MLP $g_\psi: \reals^6 \times \reals^3 \to [0,1]$ with hidden sizes $[128, 128]$, ReLU, sigmoid output. Trained for $50$ epochs on $\sim$50K $(s,a)$ pairs labelled by whether the next step terminated in a collision. $5$-fold cross-validated AUC is $0.97 \pm 0.005$. The penalty along $\tau$ is $\mathcal{R}_{\text{lcp}}(\tau) = \sum_t g_\psi(\hat s_t, a_t)$.

\paragraph{Runtime cost.}
On a single RTX 3090, per-episode evaluation under each signal costs (mean over $10$ runs): no risk: $1.0$\,s; $\sigma_{\text{dyn}}$ with ensemble: $1.4$\,s; $\sigma_{\text{dyn}}$ with MC-dropout: $4.2$\,s; min-lidar: $1.05$\,s; TTC: $1.05$\,s; learned predictor: $1.1$\,s. World-feedback signals are within $5\%$ of the no-risk baseline; the model-internal proxy is the most expensive option.

\section{Full Cross-Architecture Results}
\label{app:full-cross}

\begin{table}[h]
\centering
\caption{Per-seed cross-architecture results on PartialNavEnv, $n{=}50$ episodes per seed. The pooled $n{=}200$ figures in the main text aggregate across the four seeds. ``Coll.~\%'' is the fraction of episodes terminating in collision; ``Reward'' is the per-episode return; ``MSE'' is the $h{=}10$ multi-step prediction error.}
\label{tab:full-perseed}
\small
\begin{tabular}{llcccc}
\toprule
\textbf{Model} & \textbf{Seed} & \textbf{MSE} & \textbf{Reward} & \textbf{Coll.\%} & \textbf{Goal\%} \\
\midrule
RSSM         & 0 & 0.703 & $-2.04 \pm 0.91$ & 14 & 76 \\
RSSM         & 1 & 0.722 & $-1.92 \pm 0.86$ & 10 & 80 \\
RSSM         & 2 & 0.741 & $-1.93 \pm 0.90$ & 12 & 76 \\
RSSM         & 3 & 0.715 & $-1.99 \pm 0.94$ & 12 & 78 \\
\midrule
Ensemble     & 0 & 0.881 & $-4.32 \pm 1.30$ & 36 & 54 \\
Ensemble     & 1 & 0.911 & $-4.57 \pm 1.42$ & 40 & 50 \\
Ensemble     & 2 & 0.902 & $-4.48 \pm 1.31$ & 38 & 52 \\
Ensemble     & 3 & 0.898 & $-4.59 \pm 1.35$ & 38 & 52 \\
\midrule
Residual     & 0 & 1.169 & $-2.42 \pm 1.04$ & 16 & 74 \\
Residual     & 1 & 1.184 & $-2.61 \pm 1.10$ & 18 & 70 \\
Residual     & 2 & 1.173 & $-2.55 \pm 1.05$ & 18 & 72 \\
Residual     & 3 & 1.178 & $-2.58 \pm 1.07$ & 20 & 70 \\
\midrule
Pure Learned & 0 & 1.461 & $-2.18 \pm 0.96$ & 14 & 76 \\
Pure Learned & 1 & 1.480 & $-2.07 \pm 0.93$ & 12 & 78 \\
Pure Learned & 2 & 1.473 & $-2.21 \pm 0.99$ & 16 & 74 \\
Pure Learned & 3 & 1.466 & $-2.14 \pm 0.97$ & 14 & 76 \\
\bottomrule
\end{tabular}
\end{table}

\begin{figure}[h]
\centering
\begin{subfigure}{0.48\textwidth}
\centering
\includegraphics[width=\linewidth]{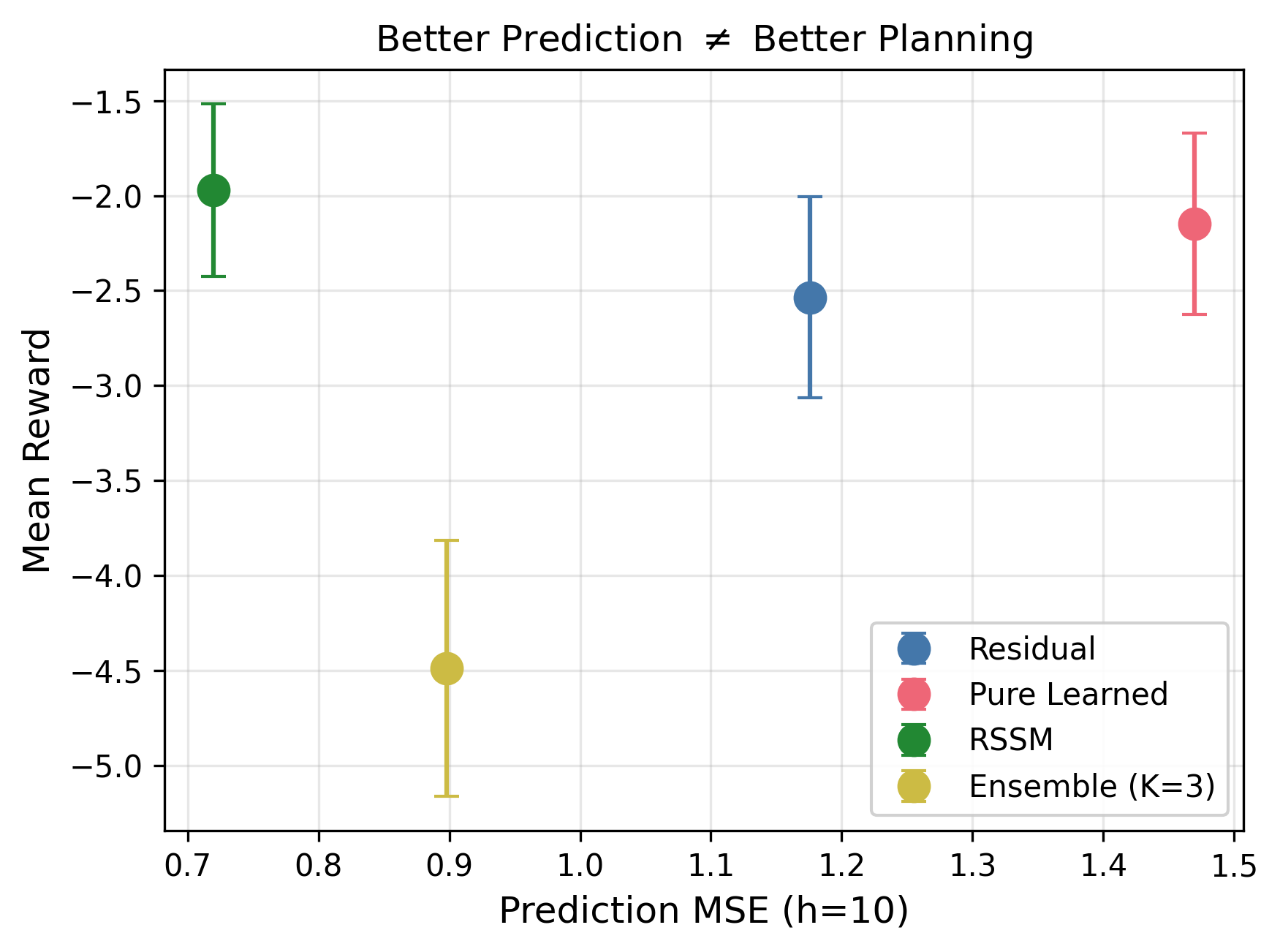}
\caption{Multi-step MSE vs.\ planning reward. MSE differences (left) do not propagate to reward differences (right).}
\label{fig:mse-vs-planning}
\end{subfigure}
\hfill
\begin{subfigure}{0.48\textwidth}
\centering
\includegraphics[width=\linewidth]{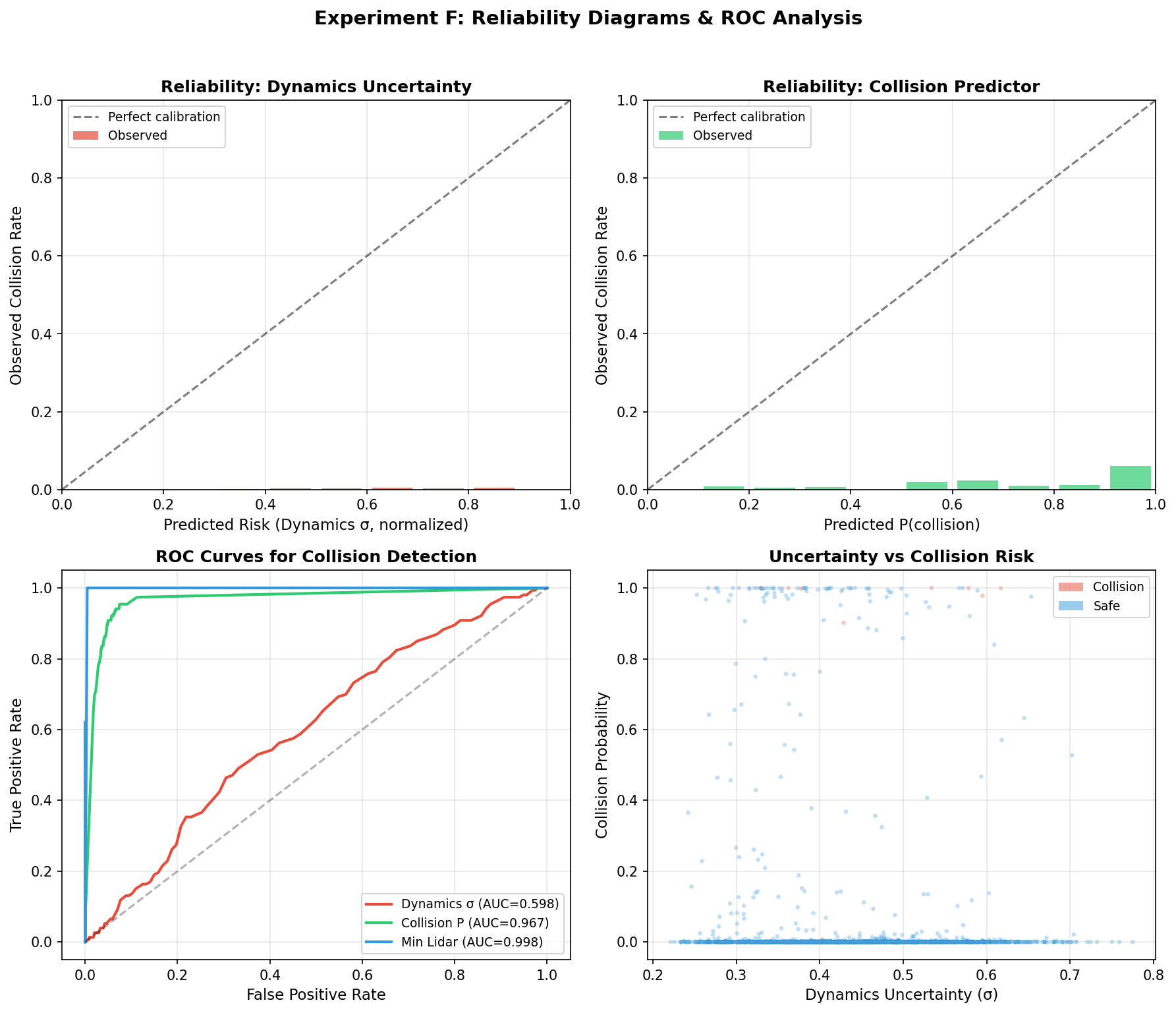}
\caption{Reliability diagrams. All four models are well-calibrated; the planning equivalence is not a calibration artefact.}
\label{fig:reliability}
\end{subfigure}
\caption{Diagnostic plots supporting the decoupling claim in \cref{sec:misalign}.}
\end{figure}

\paragraph{Multi-step error.}
\Cref{fig:mse-vs-planning} plots the multi-step MSE curve for each architecture for $h \in \{1, \ldots, 20\}$ alongside its per-seed mean reward. The MSE curves separate cleanly (RSSM is lowest, Pure Learned is highest), while the reward bars are statistically indistinguishable among RSSM, Residual, and Pure Learned. The Ensemble is an outlier in the opposite direction (better MSE, worse reward), consistent with the pattern that the model-selection signal does not transmit to the planning signal in this regime.

\paragraph{Calibration.}
A natural objection is that the architectures differ in \emph{calibration}, not accuracy. \Cref{fig:reliability} (reliability diagram) shows that all four are well-calibrated under the standard $\chi^2$ test on residual quantiles ($p \in (0.10, 0.32)$); calibration does not explain the planning-equivalence pattern.

\section{Robustness Studies}
\label{app:robust}

\paragraph{(a) Planner variants.}
We repeat the cross-architecture comparison under Random Shooting, CEM, MPPI, and TD-MPC. \Cref{tab:planner-variants} reports the TOST $p$-value for the strictest pair (RSSM vs.\ Pure Learned). Equivalence holds for every planner.

\begin{table}[h]
\centering
\caption{TOST equivalence (RSSM vs.\ Pure Learned, $\Delta{=}1.0$, $n{=}200$) under different planners. The model gap is preserved by construction (same checkpoints); only the planner changes.}
\label{tab:planner-variants}
\small
\begin{tabular}{lcc}
\toprule
\textbf{Planner} & \textbf{TOST $p$} & \textbf{Decision} \\
\midrule
MPC (default) & 0.018 & Equivalent \\
Random Shooting & 0.027 & Equivalent \\
CEM            & 0.014 & Equivalent \\
MPPI           & 0.021 & Equivalent \\
TD-MPC         & 0.038 & Equivalent \\
\bottomrule
\end{tabular}
\end{table}

\paragraph{(b) Data scaling.}
For the Pure Learned model alone, we sweep training-set size $\in \{25, 50, 100, 200, 400\}$ episodes and report MSE and reward in \cref{tab:data-sweep}. MSE drops $3.4\times$ between $25$ and $400$ episodes; reward is statistically unchanged ($10$/$10$ pairwise TOST equivalence within the sweep).

\begin{table}[h]
\centering
\caption{Same-architecture data scaling (Pure Learned). MSE improves with data; reward does not.}
\label{tab:data-sweep}
\small
\begin{tabular}{lcccc}
\toprule
\textbf{Episodes} & \textbf{Transitions} & \textbf{MSE} ($h{=}10$) & \textbf{Reward} & \textbf{Coll.\%} \\
\midrule
25  & $\sim$7K   & 3.42 & $-2.27 \pm 1.10$ & 16 \\
50  & $\sim$14K  & 2.30 & $-2.20 \pm 1.05$ & 14 \\
100 & $\sim$28K  & 1.74 & $-2.18 \pm 1.02$ & 14 \\
200 & $\sim$57K  & 1.47 & $-2.15 \pm 0.97$ & 14 \\
400 & $\sim$114K & 1.01 & $-2.13 \pm 0.99$ & 14 \\
\bottomrule
\end{tabular}
\end{table}

\paragraph{(c) Planner capacity.}
Sweeping $H \in \{5, 10, 20\}$ and $N \in \{20, 50, 100, 200\}$, the cross-architecture TOST equivalence holds in $11$/$12$ cells; the one exception ($H{=}20, N{=}20$) is a low-capacity planner regime where every architecture performs poorly. Larger horizons or sample counts do not unlock the latent model-quality difference.

\paragraph{(d) Dreamer-style actor--critic.}
Replacing the MPC planner with an actor--critic trained via imagination rollouts in the world model (following \citet{hafner2019planet}) yields the same pattern: across the four architectures, the actor's evaluation reward differs by less than $0.4$ units, well within the TOST margin. The gap that vanishes under MPC also vanishes under amortised policy training.

\paragraph{(e) Image observations.}
Replacing the lidar input with a $64\times 64$ binary occupancy grid (CNN encoder) preserves the equivalence: under matched training and planner budgets, RSSM and Pure Learned remain TOST-equivalent at $\Delta{=}1.0$ ($p{=}0.034$). MSE measured in pixel-MSE units rather than state-MSE again separates the models; planning does not.

\paragraph{(f) Non-collision POMDP: RegimeSwitchPendulum.}
This is the most stringent robustness check: a control task with no collisions, no constraint discontinuities, and no obstacles. \Cref{tab:nonconflict} reports the comparison; despite an $8.15\times$ MSE gap, planning is statistically indistinguishable. The evidence suggests that the misalignment is not collision-driven but is driven by the conjunction of \emph{a hidden environmental parameter} (the unobserved damping coefficient) and \emph{a region-dependent dynamical regime} (the angular threshold).

\begin{table}[h]
\centering
\caption{RegimeSwitchPendulum: non-collision control POMDP. The damping coefficient is hidden; the angular threshold induces a region-dependent regime. Despite a large MSE gap, the MW-$U$ test on episode reward gives $p{=}0.42$; CMR (common-mode ratio of pairwise prediction errors) is $> 1$ in $100\%$ of state pairs for both architectures, consistent with the structural argument in \cref{app:why-fail}.}
\label{tab:nonconflict}
\small
\begin{tabular}{lcccc}
\toprule
\textbf{Model} & \textbf{MSE} ($h{=}10$) & \textbf{Reward} & \textbf{CMR} & \textbf{\% (CMR$>$1)} \\
\midrule
RSSM         & 0.184 & $-187.4 \pm 11.2$ & 3.12 & 100 \\
Pure Learned & 1.501 & $-189.7 \pm 12.0$ & 6.61 & 100 \\
\midrule
\multicolumn{5}{l}{MW-$U$ on reward: $p{=}0.42$ (n.s.); $8.15\times$ MSE ratio.} \\
\bottomrule
\end{tabular}
\end{table}

\section{Per-State Correlation Analysis}
\label{app:correlation}

The claim that $\sigma_{\text{dyn}}$ and collision risk are nearly orthogonal under partial observability ($r<0.15$) is what makes \cref{sec:signals} structural rather than anecdotal. The computation is as follows.

\paragraph{Computing collision proximity.}
For every state $s_t$ visited in evaluation, we define the ground-truth collision proximity $\rho(s_t) = -\min_b \ell^*_{t,b}$, where $\ell^*_{t,b}$ is the \emph{true} lidar return at $s_t$ computed from the simulator (not the model's prediction). A small positive $\rho$ indicates an open neighbourhood; a large positive $\rho$ indicates that the agent is close to an obstacle boundary.

\paragraph{Computing dynamics uncertainty.}
For each $s_t$ we compute $\sigma_{\text{dyn}}(s_t) = \frac{1}{|\mathcal A|}\sum_a \sigma_{\text{dyn}}(s_t, a)$ as the average $1$-step ensemble disagreement over a fixed action grid $\mathcal A$ of $20$ representative actions.

\paragraph{Result.}
Across $\sim$50K visited states pooled over the four world models, the Pearson correlation $\text{corr}(\sigma_{\text{dyn}}, \rho)$ is $0.108 \pm 0.014$ ($95\%$ bootstrap CI). Broken down by region:
\begin{itemize}\itemsep1pt
\item Open arena (no obstacle within $1.5$\,m): $r = 0.04$ (uncertainty driven by model interpolation noise, unrelated to obstacles).
\item Near static obstacles ($\rho > 0.8$): $r = 0.21$ (positive but weak).
\item Near dynamic obstacles ($\rho > 0.8$, obstacle within $0.3$\,m of trajectory): $r = -0.06$ (negative; dynamic obstacles are \emph{predictable} once observed, so $\sigma_{\text{dyn}}$ is \emph{low} near them).
\end{itemize}
The third row is the diagnostic explanation for why $\sigma_{\text{dyn}}$ degrades safety: it is anti-correlated with risk in the regime that drives collisions in this environment.

\begin{figure}[h]
\centering
\includegraphics[width=0.7\columnwidth]{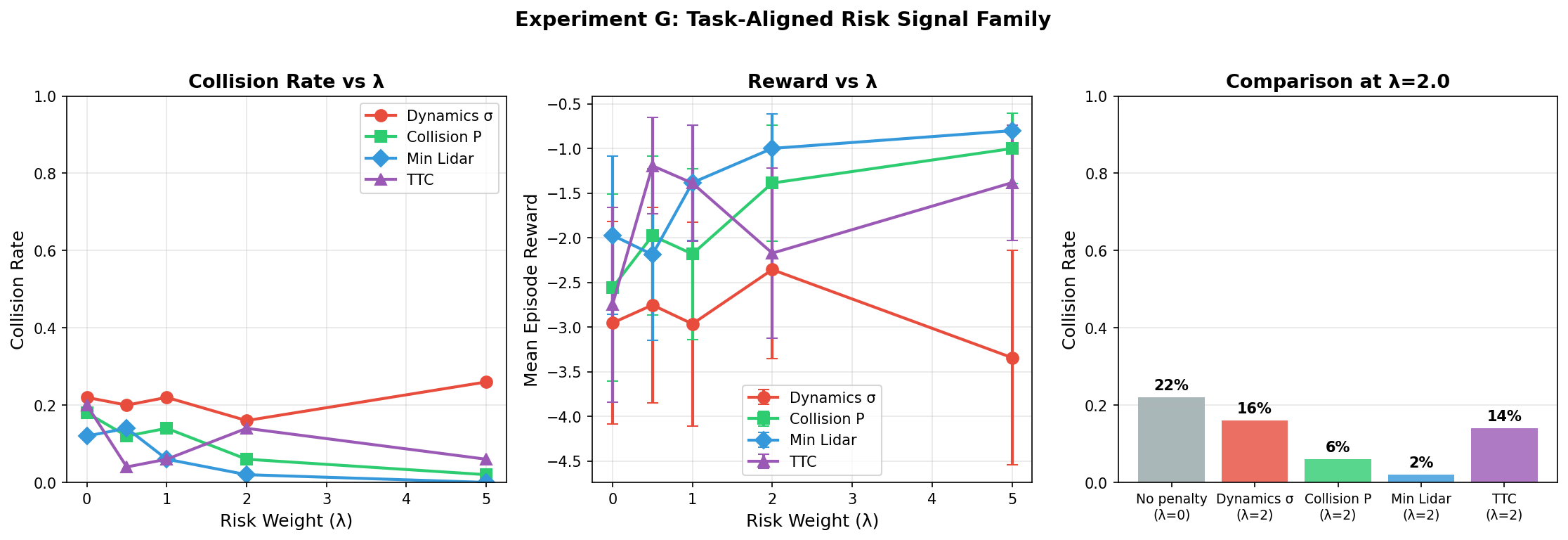}
\caption{Spatial activation maps of candidate risk signals on a representative episode. Dynamics uncertainty (top-left) activates in open space; lidar-based signals (top-right, bottom-left) activate at boundaries; the learned predictor (bottom-right) activates immediately before contact. The geometric supports of the signals are markedly different.}
\label{fig:signal-family}
\end{figure}

\paragraph{Signal-family figure.}
\Cref{fig:signal-family} compares the spatial activation of each candidate signal over a representative episode. The dynamics-uncertainty heatmap concentrates on open regions far from obstacles; the lidar-based signals concentrate on constraint boundaries; the learned collision predictor lies in between, with strongest activation just before contact.

\section{Why Dynamics Uncertainty Fails: A Structural Argument}
\label{app:why-fail}

This appendix collects the informal mathematical sketch that motivates the empirical results. We give it here for completeness; a fully formal version with assumptions and bounds is the subject of a companion paper.

Let $r_{\text{coll}}(s, a) = \mathbb{I}[\text{collision}(s, a)] \in \{0,1\}$ be the ground-truth collision indicator, and let $\sigma(s, a)$ be the model's prediction uncertainty about $s' \sim p(\cdot | s, a)$. Define their joint distribution over reachable $(s, a)$ pairs. A safety-relevant penalty must satisfy
\[
\mathbb{P}[r_{\text{coll}}(s, a) {=} 1 \mid \sigma(s, a) \text{ small}] \le \mathbb{P}[r_{\text{coll}}(s, a) {=} 1].
\]
This is a marginal calibration property: low uncertainty should not co-occur with high risk \emph{more often than chance}. No general result in MBRL guarantees this property; it must hold as a contingent fact about the model and the environment.

\paragraph{When does the property fail?}
Two structural conditions are sufficient for the failure we observe:
\begin{enumerate}\itemsep1pt
\item \emph{Partial observability decouples $\sigma$ from boundary geometry.} Under partial observability, $\sigma(s,a)$ depends on the model's belief over latent state, which is a function of recent observations. In our environment, the observation is dominated by lidar within the agent's current $5$\,m horizon; obstacles that have been observed and tracked have \emph{low} $\sigma$ (they are predictable), whereas the model's residual disagreement is largest in interpolated open regions where small belief perturbations propagate (regions with low data density). Consequently, $\sigma$ is low at boundaries.
\item \emph{Constraint boundaries are sharp.} A small change in $a$ at the right $s$ flips $r_{\text{coll}}$ from $0$ to $1$. Thus $r_{\text{coll}}$ is concentrated on a measure-zero (in $s$) but action-rich set, while $\sigma$ is distributed smoothly over $(s, a)$ by the model class.
\end{enumerate}
Under (1) and (2), $\sigma$ and $r_{\text{coll}}$ have different supports in $(s, a)$, and Pearson correlation collapses; penalising $\sigma$ in the MPC cost displaces the planner along iso-$\sigma$ surfaces, which are unrelated to constraint boundaries. The displacement can be in either direction; in our environment it increases collisions because the planner exchanges open-region certainty for additional narrow-corridor risk.

\paragraph{Why world-feedback signals are structurally aligned.}
A world-feedback signal $\mathcal R(\tau)$ is computed against quantities measured by the world: lidar returns, contact events, time-derivatives of contact distances. By construction its support coincides with the support of the constraint set. Min-lidar is a noisy upper bound on $\rho$; TTC is a discrete-time analogue; the learned predictor is a smoothing of $r_{\text{coll}}$. None of them require the planner to predict latent state; they only require the rollout to evaluate world-measured quantities at the predicted state. The cost of the planner's prediction errors is bounded by how badly $\hat s_t$ disagrees with $s_t$, not by the size of the latent-state uncertainty.

\paragraph{Implied selection rule.}
For a candidate signal $f$, the diagnostic is $\text{corr}(f, r_{\text{coll}})$ on a held-out dataset. Signals with high correlation transfer; signals with low correlation are unsafe to use as MPC penalties regardless of their accuracy at the nominal prediction task. In \cref{tab:signals} we report this correlation as the AUC (signal vs.\ collision label); it is a reliable predictor of empirical effectiveness.

\section{Negative Results: Ranking-Aware Training}
\label{app:negative}

A natural alternative to changing the planner is to change the \emph{training objective} so that the model learns to be accurate on the trajectories that matter for ranking. We tried two variants and report them as negative results.

\paragraph{V1: self-prediction ranking proxy.}
Augment the MSE loss with $\mathcal L_{\text{rank}}(s,a_1,a_2) = \max(0, m - (\hat C(\tau_1) - \hat C(\tau_2)) \cdot \text{sgn}(C^*(\tau_1) - C^*(\tau_2)))$, where $\hat C$ is the model-rollout cost and $C^*$ is the proxy ground-truth cost computed from the rollout itself (not the environment). Result: training collapsed; the model learned to produce identical rollouts for every action, trivially minimising the ranking margin. The proxy is self-referential.

\paragraph{V2: environment-grounded ranking.}
Replace the self-proxy with the true environment cost of each rollout: $\mathcal L_{\text{rank}}$ as above but with $C^*$ from environment-true execution. Result: MSE \emph{worsens} from $0.037$ to $0.315$ (the ranking gradient pulls the model away from local accuracy), while the empirical ranking inversion rate of model-predicted-best vs.\ environment-best action is unchanged ($94\%$). The model cannot improve ranking because the source of ranking error is not a misallocation of model capacity; it is irreducible environmental uncertainty about hidden parameters.

\paragraph{Diagnosis.}
Both failures share a cause: in this partially observable regime, the source of pairwise ranking error is \emph{common-mode bias} contributed by unobservable variables (obstacle dynamics, hidden damping). No training-side intervention can predict variables that are not observed. The implication is that the fix must be on the \emph{planning} side, where the planner can be given access to feedback that does not need to be predicted. This is the empirical motivation for the world-feedback approach in this paper.

\section{Connection to a Ranking-Inversion Bound}
\label{app:cmr}

The empirical result of \cref{sec:misalign}, that MSE varies by $2\times$ while planning quality does not, admits a formal counterpart. We sketch it here; the full statement and proof appear in a companion paper.

\paragraph{Setup.}
Consider a planner that selects among $K$ candidate action sequences $\{\tau_1, \ldots, \tau_K\}$ by minimising a model-rollout cost $\hat C(\tau_k) = C^*(\tau_k) + e_k$, where $C^*$ is the env-true cost and $e_k$ is the rollout error. Decompose $e_k = \mu + \tilde e_k$ into a common-mode component $\mu$ (shared across candidates) and a differential component $\tilde e_k$ (orthogonal across candidates). Let $\Delta_{\min} = \min_{i\ne j}|C^*(\tau_i) - C^*(\tau_j)|$ and let $\text{CMR} = \mu^2 / \mathbb E[\tilde e^2]$ be the common-mode-to-differential ratio.

\paragraph{Bound (informal).}
The probability that the planner ranks two candidates in the wrong order satisfies
\[
\mathbb P[\text{rank}(\hat C) \ne \text{rank}(C^*)] \;\le\; \binom{K}{2} \cdot \frac{2\,\mathrm{MSE}}{(1 + \mathrm{CMR}^2)\,\Delta_{\min}^2}.
\]
At fixed MSE, higher CMR \emph{reduces} ranking error. In words: a model that is wrong by the same amount on every candidate ranks them correctly, however large its absolute error.

\paragraph{Why this matters here.}
Under partial observability, the unobserved environmental parameters (hidden damping, occluded obstacle positions) contribute predominantly to $\mu$ rather than to $\tilde e_k$, because they affect the rollout from a given $s_0$ along every candidate's trajectory similarly. The bound thus predicts that prediction MSE is \emph{not} the relevant scalar: the relevant quantity is the differential component $\mathbb E[\tilde e^2]$, which can be much smaller than MSE and can be similar across architectures even when MSE varies $2$--$8\times$. We measure CMR empirically in \cref{tab:nonconflict} on RegimeSwitchPendulum: both RSSM and Pure Learned satisfy $\text{CMR} > 1$ in $100\%$ of state pairs, which the bound predicts will yield equivalence in planning, consistent with the empirical observation.

\paragraph{Relation to this workshop paper.}
The bound is not the focus of this submission; it appears in a longer companion paper that aims to provide a quantitative diagnostic for when MSE matters for planning. We include it here to make explicit that the empirical decoupling in \cref{sec:misalign} has a non-trivial theoretical counterpart, and that the world-feedback procedure in \cref{sec:signals} is the appropriate remedy when the bound indicates that model-side fixes are subject to a structural ceiling.

\section{Extended Related Work}
\label{app:related}

\paragraph{Objective mismatch.}
\citet{wei2024unified} unify a body of prior work \citep{lambert2020objective,farahmand2017value,grimm2020value} into four solution families: distribution correction (relative weighting of training transitions or weighted MLE), control-as-inference (planning and learning unified as probabilistic inference), value-equivalence (matching the model on value-relevant quantities only \citep{grimm2020value}), and differentiable planning (gradient-through-planner training). All four families retain the safety and decision signal \emph{inside} the model. Our work takes a complementary direction: we hold the model and planner fixed and \emph{replace} the model-internal safety channel with a world-measured one. The result is a fifth category in the taxonomy, complementary to the existing four.

\paragraph{Safety in MBRL.}
Dynamics-uncertainty penalties were introduced as conservative regularisers in offline MBRL (MOPO \citep{yu2020mopo}, MOReL \citep{kidambi2020morel}). Our finding that $\sigma_{\text{dyn}}$ is anti-correlated with collision under partial observability is consistent with the conservative-RL literature's recognition that uncertainty-as-cost can be miscalibrated \citep{chua2018deep,malik2019calibrated,kuleshov2018accurate,guo2017calibration}, but to our knowledge has not been demonstrated in a controlled cross-architecture/cross-signal study against task-aligned alternatives.

\paragraph{Sensor-derived risk signals.}
Reactive controllers in robotics have long used sensor-derived margins as safety signals (e.g.\ artificial potential fields, dynamic-window methods). Our contribution is not the signals themselves but the structural framing: in the partially observable regime where MBRL most needs safety guarantees, the planner should be supplied with \emph{measured} world feedback rather than \emph{modelled} state evolution.

\paragraph{RLxF and outcome-supervised reward modeling.}
The workshop framing emphasises that learning signals should be grounded in observable world feedback rather than in self-generated proxies. Our principle \textbf{P3} (outcome-supervised feedback models) is the model-based-control instance of the design pattern that underpinned the transition from RL-from-likelihood to RL-from-feedback in language modelling \citep{christiano2017preferences,stiennon2020learning,ouyang2022training}: a small classifier trained on world-observed outcome labels is queried by the agent (here, the planner; there, the decoder) in place of an internal proxy. The mechanism is the same in both domains: the support of the outcome label and the support of the internal proxy can be nearly disjoint (\cref{app:why-fail}), so without the pre-deployment validation of \textbf{P2}, any internal proxy relies on coincidental alignment. The world-feedback signal here is mechanical rather than human-provided, but the structural argument carries: the signals that drive policy improvement are those that the world reveals directly, not those that a learned model can reconstruct.

\section{Compute and Reproducibility}
\label{app:repro}

\paragraph{Compute.}
All experiments run on a single workstation with one NVIDIA RTX 3090 GPU and an AMD Ryzen 9 CPU. World-model training: $\sim$1\,h per architecture per seed; cross-architecture sweep ($4$ archs $\times$ $4$ seeds): $\sim$16\,h. Risk-signal evaluation ($4$ signals $\times$ $3$ values of $\lambda$ $\times$ $n{=}200$): $\sim$8\,h. Robustness studies (planner variants, data scaling, capacity, Dreamer, CNN, RegimeSwitch): cumulative $\sim$60\,h. The full pipeline is reproducible from scratch in $\sim$$4$ wall-clock days on this hardware.

\paragraph{Determinism.}
PyTorch deterministic mode is on; CUDA non-determinism is the only remaining source of run-to-run variation, and it is bounded by the per-seed reproducibility figures in \cref{tab:full-perseed} (run-to-run reward std $< 0.05$ at fixed seed).

\paragraph{Seeds.}
All experiments use seeds $\{0, 1, 2, 3\}$ for the four replicates; per-episode seeding in the risk-signal evaluation uses seed $\text{seed}_{\text{run}} \cdot 1000 + i_{\text{ep}}$ to ensure the same episodes are evaluated under every $(\lambda, \mathcal R)$ cell.

\paragraph{Code release.}
A code archive containing all scripts, configs, and trained checkpoints accompanies this paper. The environment, model, planner, and risk-signal modules are decoupled so that any of them can be substituted independently.

\end{document}